\documentclass[10pt,twocolumn,letterpaper]{article}

\usepackage{cvpr}
\usepackage{times}
\usepackage{epsfig}
\usepackage{graphicx}
\usepackage{amsmath}
\usepackage{amssymb}
\usepackage{footnote}
\usepackage{multirow}
\usepackage{tabularx}
\usepackage{paralist}
\usepackage{booktabs}

\usepackage[caption=false,font=footnotesize,labelfont=sf,textfont=sf,subrefformat=parens,labelformat=parens]{subfig}
\usepackage[caption=false,font=footnotesize,subrefformat=parens,labelformat=parens]{subfig}

\usepackage[pdfstartview=FitH,pagebackref=true,breaklinks=true,colorlinks,bookmarks=false,linkcolor=blue,citecolor=blue]{hyperref}

\cvprfinalcopy 

\def\cvprPaperID{2877} 

\ifcvprfinal\fi
\begin{document}

\title{Generative Single Image Reflection Separation}

\author{Donghoon Lee$^1$, Ming-Hsuan Yang$^2$, and Songhwai Oh$^1$\\
$^1$Electrical and Computer Engineering and ASRI, Seoul National University, Korea\\
$^2$Electrical Engineering and Computer Science, University of California at Merced\\
{\tt\small donghoon.lee@cpslab.snu.ac.kr, mhyang@ucmerced.edu, songhwai@snu.ac.kr}}

\maketitle

\begin{abstract}
Single image reflection separation is an ill-posed problem since two scenes,
a transmitted scene and a reflected scene, need to be inferred from a single observation.
To make the problem tractable,
in this work we assume that categories of two scenes are known.
It allows us to address the problem by generating both scenes that belong to the categories
while their contents are constrained to match with the observed image.
A novel network architecture is proposed to render realistic images of both scenes based on adversarial learning.
The network can be trained in a weakly supervised manner,
i.e., it learns to separate an observed image
without corresponding ground truth images of transmission and reflection scenes
which are difficult to collect in practice.
Experimental results on real and synthetic datasets demonstrate that the proposed algorithm
performs favorably against existing methods.
\end{abstract}

\vspace{-2mm}
\section{Introduction}
\vspace{-2mm}
Single image reflection separation aims to separate an observed image into a transmitted scene and a reflected scene.
When two scenes are separated adequately,
existing computer vision algorithms can better understand each scene
since an interference of the other one is decreased.
As there are various objects that may reflect surroundings,
such as windows, glass, or ponding water,
this is an important problem to the computer vision community.

\begin{figure}[t]
    \centering
    \includegraphics[width=\linewidth]{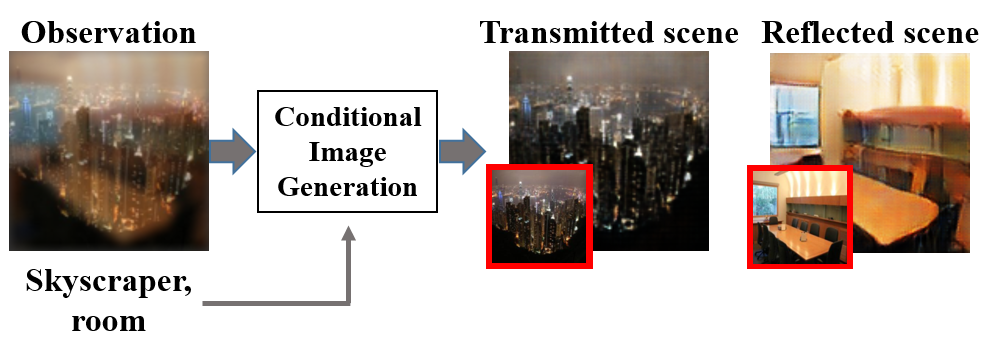}
    \caption{
    The proposed algorithm is able to separate an observed image into a transmitted scene and a reflected scene.
    Convolutional neural networks are trained to generate both scenes conditioned on the category of two scenes
    based on adversarial learning.
    Unlike previous approaches, a reflected scene can also be predicted reasonably.
    The ground truth of each scene is shown in red boxes.
    }
\label{fig:teaser}
\end{figure}

Although this problem has been studied for decades \cite{barrow1978computer}, it is still a challenging problem due to several reasons.
First, it is an ill-posed problem as we need to infer two scenes based on an observed image.
Numerous methods have been proposed to address this issue by making certain assumptions to make the problem tractable.
However, the assumptions have been limited to specific cases and are not applicable to real-world images in general \cite{wan2017benchmarking}.
For example, one of the mainstream approaches assumes that the edge distribution between the transmitted scene and reflected scene is different, i.e., the former tends to have sharper edges while the latter is relatively blurred \cite{arvanitopoulos2017single, li2014single, fan2017generic}.
This type of blur occurs when the reflected scene is outside of the depth of field of the camera.
However, cameras in recent years, such as the ones on smartphones, have small aperture and deep depth of field.
Consequently, an observed image contains sharp edges of both scenes although the reflected scene is not in the same depth of the transmitted scene.
In other words, the blur assumption does not hold in practice.

Second, it is difficult to obtain ground-truth tuples of an observed scene $y$, a transmitted scene $t$, and a reflected scene $r$ as we cannot simply get rid of the obstructing glass and take pictures of two scenes in general circumstances\footnote{Note that a transmitted scene and a reflected scene represent a scene before the transmission and reflection, respectively.}.
Therefore, it has been assumed that an observed image can be synthesized by combining two known images based on the physical model of the transmission and reflection.
However, the exact physical model is unknown as it is complex and requires various factors, such as the thickness of glass, surface conditions, and angle of incidence, which are typically not at our disposal \cite{soares2014introduction}.
As a consequence, simple approximations are used to model the reflection which limits the separation performance.
For example, one of the most common approximated models is $y=wt+(1-w)r$, where $w$ is a scalar weight.
However, it does not consider changes in $t$ and $r$ due to the glass.

Third, conventional approaches mainly focus on reflection removal instead of reflection separation~\cite{han2017reflection, arvanitopoulos2017single, wan2016depth, xue2015computational}.
Although these methods aim to suppress reflection artifacts to restore the transmitted scene,
the contents of the reflection are usually not considered.
While it may be difficult to reconstruct both scenes,
it is important to infer both scenes jointly as they are entangled in a single observation.
Furthermore, the recovered reflected scene itself may be useful for various applications such as surveillance and image understanding.

To address above issues, we first propose to use a new assumption
on the observed scene: the category of the transmitted and reflected scenes are known.
%
%
%
It is a valid assumption in practice since we take a picture of an interested object or scene while knowing which is being reflected.
Likewise, other ill-posed problems such as deblurring and super-resolution often make similar assumptions, e.g., in the cases of faces, text, and night scenes~\cite{pan2014deblurring, pan2014deblurring2, hu2014deblurring, yang2013structured}.
As such, we demonstrate that it is not necessary to make other assumptions, e.g., a blurry reflected scene.
In addition, our algorithm does not rely on a certain approximation model of the reflection as it may restrict the algorithm to a specific case.
Instead, we leverage the fact that an observed image contains contents of the transmitted scene and reflected scene.
It leads us to model an observed image using a feature space instead of a pixel-level combination.

Based on the above assumption, we pose the reflection separation problem as a conditional image generation task (see Figure~\ref{fig:teaser}).
Although image generation is a difficult problem, notable success has been achieved that transforms an input image into other domains or styles \cite{isola2017image, gatys2016image, zhu2017unpaired}.
In this work, we use generative adversarial networks (GAN) \cite{goodfellow2014generative}
to infer two scenes jointly based on
a novel network architecture.
By generating images, the proposed algorithm makes a key difference from previous methods of reflection separation as plausible reflected scenes can be obtained.
Furthermore, the network can be trained in a weakly-supervised manner, i.e., only
labels of transmitted and reflected scenes are needed.
It enables us to train the network using real data without tedious effort gathering corresponding ground truths for $t$ and $r$.

We carry out experiments on real data collected from the internet and synthetic data based on the Places dataset \cite{zhou2017places} which consists of 8 million images of 365 scenes.
For both datasets, we evaluate the proposed algorithm against the state-of-the-art methods for single image reflection separation.
Quantitative and qualitative results show that the proposed algorithm
suppresses reflection artifacts on the transmitted scene
and infers the reflected scene properly.

\vspace{-2mm}
\section{Related Work}
\vspace{-2mm}
As reflection separation is an ill-posed problem,
additional information or assumptions are needed to make the problem tractable.
In earlier methods, multiple images of a target scene taken under different conditions are used.
For example,
focus/defocus pairs \cite{schechner2000separation},
flash/non-flash pairs \cite{agrawal2005removing},
or different polarization angle pairs \cite{schechner2000polarization, kong2014physically}
are utilized.
For videos, it is possible to decorrelate the motion between the transmitted scene and reflected scene \cite{sarel2004separating, szeliski2000layer, gai2012blind}.
However, it may be difficult to apply these methods in practice since multiple images captured from controlled experimental setups are not always available.

Reflection separation using a single image
has recently attracted increasing attention due to its practical importance,
although the problem is more difficult than multiple-image cases.
User annotations can guide the separation by formulating it as a constrained optimization problem which relies on a sparse gradient prior of natural images \cite{levin2007user}.

For automatic single image reflection separation,
existing methods focus on each of the following three different conditions.
First, the depth of field (DoF) of a lens is shallow and the image is focused on a transmitted scene.
It makes out-of-focus blur for a reflected scene when the distance from the window is not the same as the transmitted scene.
Thus, blurry edges become useful cues for reflection separation.
Wan \etal~\cite{wan2016depth} propose a pixel-wise DoF confidence map obtained by a multi-scale search.
In \cite{yan2014separation}, a method based on a Markov random field and expectation maximization is proposed to filter out weak edges.
The energy function of a Markov random field is composed of gradient profile sharpness and spatial smoothness of edges.
A recent optimization based approach \cite{arvanitopoulos2017single} uses a Laplacian data fidelity term and an $l_0$ prior term to suppress reflections.
Fan \etal~\cite{fan2017generic} propose a two-step deep architecture based on convolutional neural networks.
Given an input image and an edge map,
it first predicts edges of a target scene
and then reconstructs the target scene based on the input image and predicted edges.

\begin{figure*}[!t]
    \captionsetup[subfloat]{farskip=2pt,captionskip=1pt}
    \centering
    \subfloat[Baseline 1]{\includegraphics[width=0.25\linewidth]{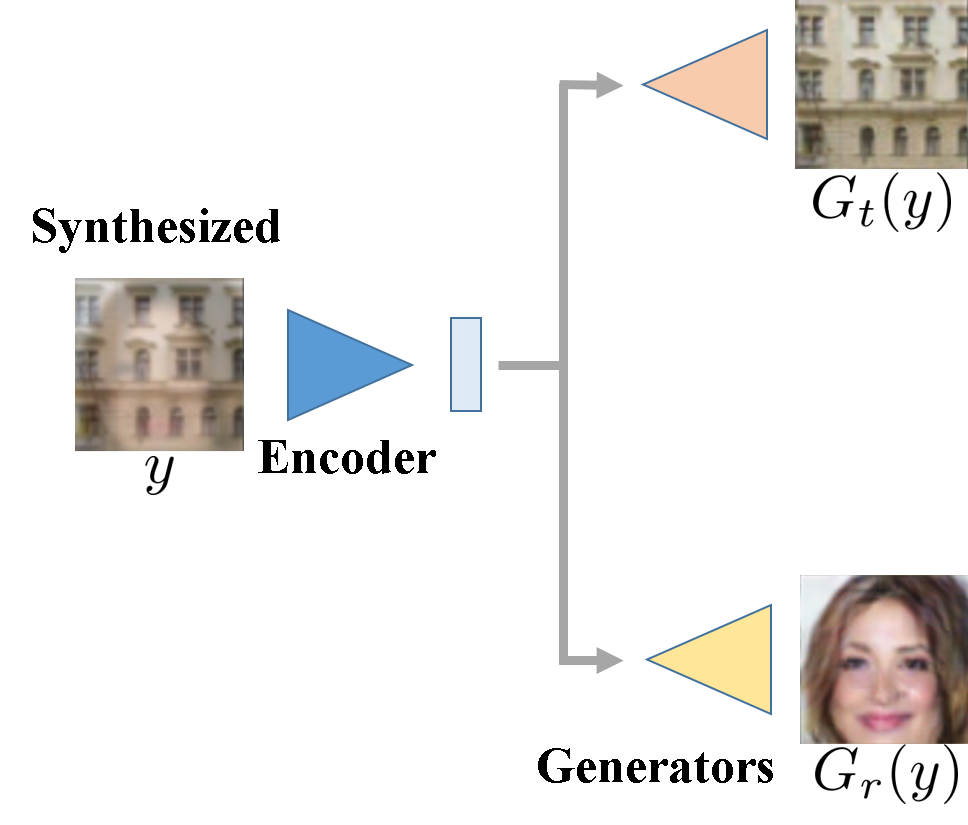}%
    \label{fig:net1}}
    \qquad
    \subfloat[Baseline 2]{\includegraphics[width=0.25\linewidth]{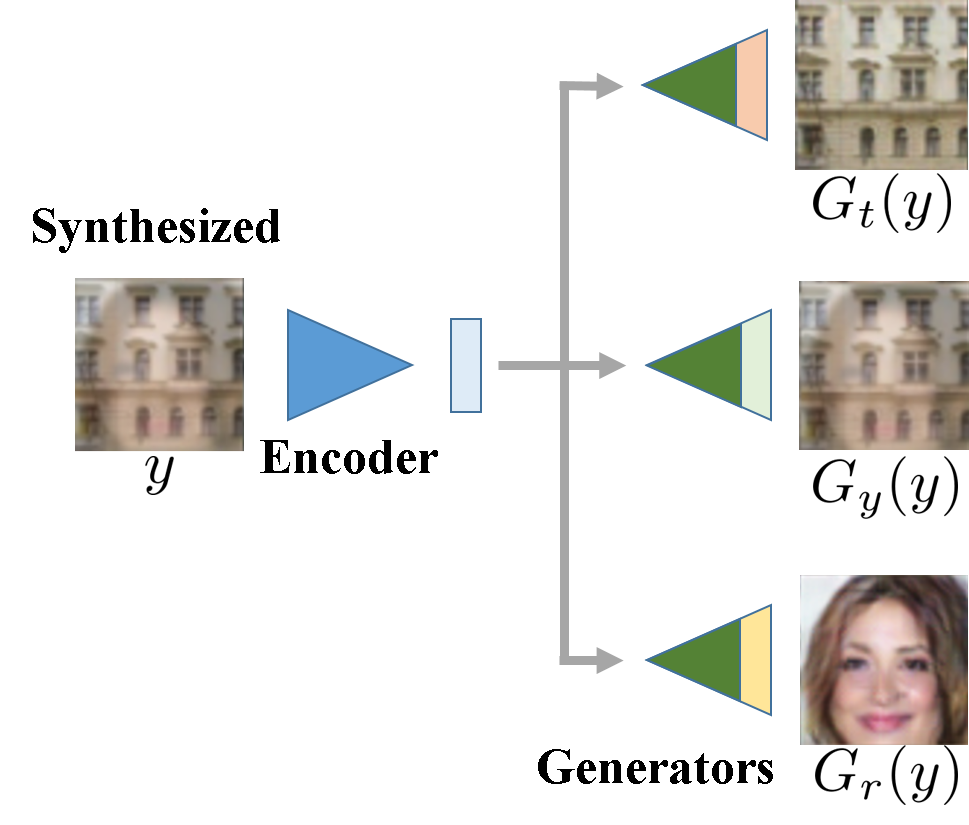}%
    \label{fig:net2}}
    \qquad
    \subfloat[Baseline 3]{\includegraphics[width=0.315\linewidth]{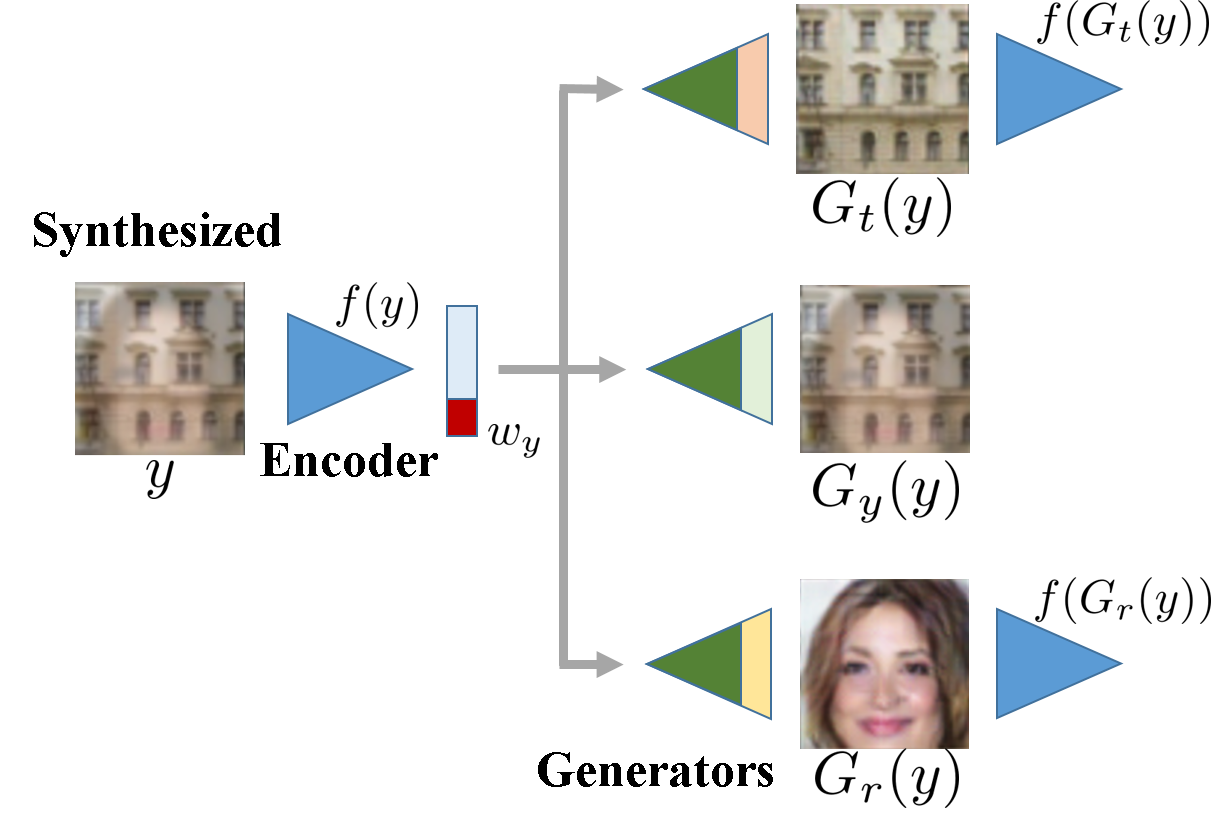}%
    \label{fig:net3}}
    \hfil
    \subfloat[Proposed network]{\includegraphics[width=0.6\linewidth]{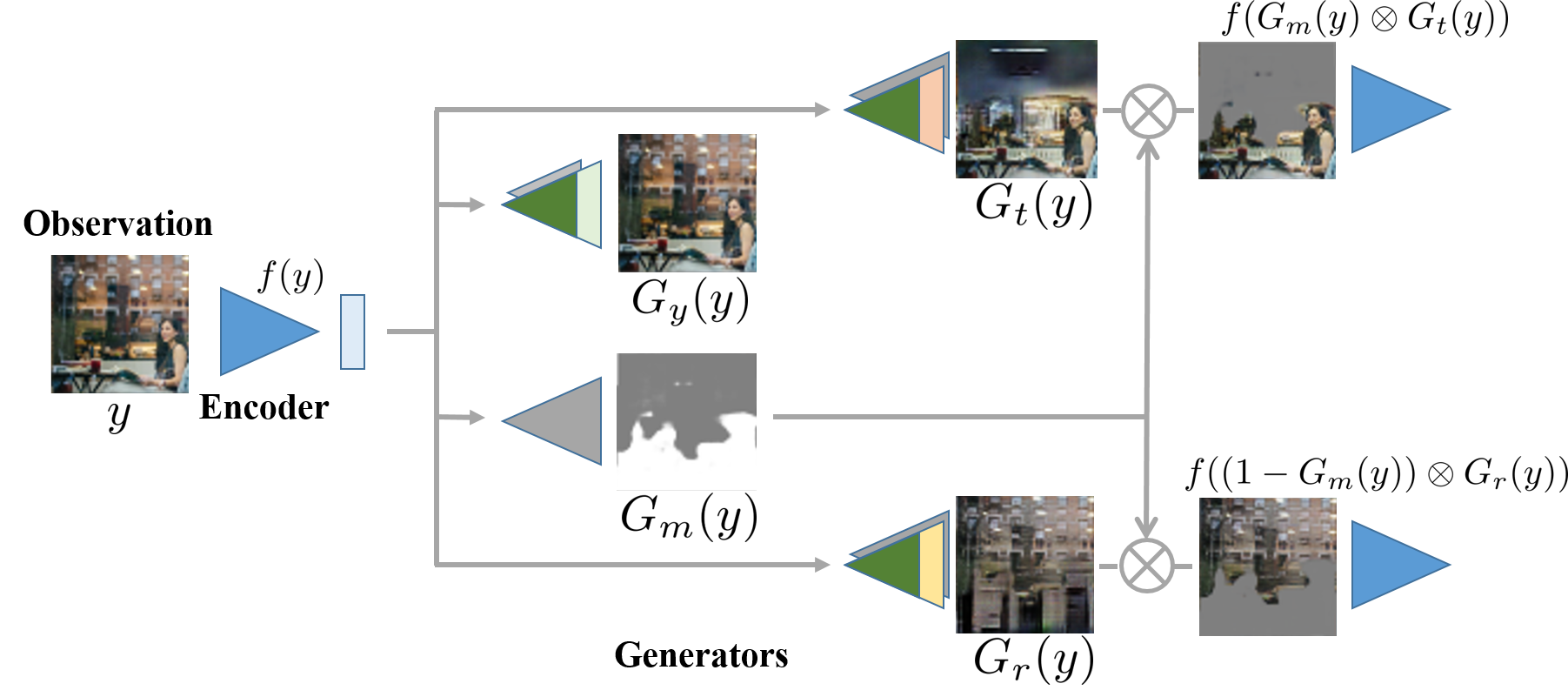}%
    \label{fig:net4}}
    \hfil
    \caption{
    Proposed network architecture and baseline models for reflection separation.
    We use color coding to indicate weight sharing.
    Note that U-net architecture and discriminators are omitted for better visualization.
    (a) Two separated generators for each of transmitted and reflected scenes.
    (b) Sharing weights through a reconstruction branch.
    (c) Additionally comparing contents of images using feature maps of a shared encoder.
    (d) Predicting a mask map for a transmitted scene increases the flexibility of a network for handling real images.
    }
    \label{fig:nets}
\end{figure*}

Second, the DoF is deep and covers both transmitted and reflected scenes.
Thus, the observed image contains sharp edges from both scenes.
While this happens frequently, it is the most challenging issue due to the lack of visual clues for separation.
Levin \etal~\cite{levin2003learning, levin2004separating} rely on a prior that gradient and local features of natural images are statistically sparse~\cite{fergus2006removing}.
However, it is difficult to apply above methods when textures of an image become complex.

Third, the DoF is deep and glass is relatively thick.
Then, a ghost effect of the reflected scene is caused by light rays that penetrate the outer surface of the glass but reflect on the inner surface.
Shih \etal~\cite{shih2015reflection} estimate a ghost kernel and use a Gaussian mixture model to separate the scenes.
However, it is not applicable to other types of reflections that do not have notable ghost effects.

\vspace{-2mm}
\section{Proposed Algorithm}
\vspace{-2mm}
\begin{table*}[t]
    \centering
    \footnotesize
    \caption{Details of each network. \# Filter is the number of filters. BN is the batch normalization. Conv denotes a convolutional layer. F-Conv denotes a transposed convolutional layer that uses the fractional-stride.}
    \vspace{-2mm}
\subfloat[Details of the \{encoding, discriminator\} network]{
    \begin{tabular}{|c|c|c|c|c|c|}
    \hline
    Layer & \# Filter & Filter Size & Stride & Pad & BN \\
    \hline
    Conv. 1 & 32   & $5\times5\times3$   & 2 & 2 & $\times$ \\
    Conv. 2 & 64  & $5\times5\times32$  & 2 & 2 & \{$\times$, $\bigcirc$\} \\
    Conv. 3 & 128  & $5\times5\times64$ & 2 & 2 & \{$\times$, $\bigcirc$\} \\
    Conv. 4 & 256  & $5\times5\times128$ & 2 & 2 & \{$\times$, $\bigcirc$\} \\
    Conv. 5 & 256 & $5\times5\times256$ & 2 & 2 & \{$\times$, $\bigcirc$\} \\
    Conv. 6 & \{128,1\} & $1\times1\times256$ & 1 & 0 & $\times$ \\
    \hline
    \end{tabular}%
  \label{tab:EnDi}%
}
\subfloat[Details of the \{mask prediction, generation\} network]{
    \begin{tabular}{|c|c|c|c|c|c|}
    \hline
    Layer & \# Filter & Filter Size & Stride & Pad & BN \\
    \hline
    Conv. 1 & $4\times4\times256$ & $1\times1\times128$ & 1 & 0 & $\bigcirc$ \\
    F-Conv. 2 & 256 & $5\times5\times256$     & 1/2 & - & $\bigcirc$ \\
    F-Conv. 3 & 256  & $5\times5\times128$     & 1/2 & - & $\bigcirc$ \\
    F-Conv. 4 & 128  & $5\times5\times64$     & 1/2 & - & $\bigcirc$ \\
    F-Conv. 5 & 64  & $5\times5\times32$      & 1/2 & - & $\bigcirc$ \\
    F-Conv. 6 & 32   & $5\times5\times\{1,3\}$ & 1/2 & - & $\times$ \\
    \hline
    \end{tabular}%
  \label{tab:DeGe}%
}
\label{tab:layers}
\end{table*}

Figure~\ref{fig:nets} shows design options of network architectures for a single image reflection separation.
From baseline models to the proposed network, we discuss the limitations and remedies of each model.
All networks basically have two generation branches; one for a transmitted scene and the other for a reflected scene.
For realistic generations, we apply adversarial losses~\cite{goodfellow2014generative} using discriminators.
Let $(G_t, D_t)$ and $(G_r, D_r)$ denote pairs of a generator and a discriminator for transmission and reflection branches, respectively.
Then, an adversarial loss for the transmission branch is defined as follows:
\begin{equation}
\begin{aligned}
    \mathcal{L}_{adv}(G_t,D_t) =
    & \mathbb{E}_{y,t\sim p_{data}(y,t)}[\log D_t(y,t)] + \\
    & \mathbb{E}_{y\sim p_{data}(y)}[\log (1-D_t(G_t(y))].
\end{aligned}
\label{eq:cgan_loss}
\end{equation}
An adversarial loss for the reflection branch is defined in a similar way.
The network architecture is described in Table~\ref{tab:layers}.

\subsection{Reflection Separation using Synthetic Images}
\vspace{-2mm}
This approach trains a network to separate synthesized images and evaluate on real images.
We first describe our synthesis schemes and then discuss network designs.

\vspace{2pt}
\noindent {\bf Preparing training images.}
For training, $y$ is synthesized using two known images $t$ and $r$, and a synthesis model.
We consider three state-of-the-art synthesis models \cite{arvanitopoulos2017single, shih2015reflection, fan2017generic}:
\begin{compactitem}
  \item [\textbf{A blurring model}] $y=wt+(1-w)(k_b*r)$,
  \item [\textbf{A ghost model}] $y=t+k_g*r$,
  \item [\textbf{A clipping model}] \cite{fan2017generic},
\end{compactitem}
where $w$ is a scalar weight, $k_b$ is a blurring kernel, $k_g$ is a ghost kernel, and $*$ denotes a convolution operation.
These models assume blurry edges or ghost effects of a reflected scene to make the problem tractable.
We aim to drop such assumptions and
thus two more synthesis models are added, i.e., a linear model $y=wt+(1-w)r$ and a clipping model without blurring step.
%
In addition, we change the ghost model as $y=wt+(1-w)\frac{k_g*r}{\max(k_g*r)}$ for training since the model in \cite{shih2015reflection} does not define $t$ and $r$ as the original scenes before the transmission and reflection\footnote{For the original ghost model, $y$ is outside of $[0,1]$ when $t\in[0,1]$ and $r\in[0,1]$ since $k_g*r\geq r$.}.

There are parameters in each model, \eg, $w$ in the blurring model.
For flexibility and generalization ability, we synthesize images using random parameters for each mini-batch.
For example, we pick $w\in[0.5,0.7]$ uniformly at random while other methods use one or two fixed values \cite{arvanitopoulos2017single, li2014single}.
In addition, we perform random left-right flipping and cropping for data augmentation.
To the best of our knowledge, this is the first work that deals with these diverse reflection models.

\begin{figure}[t]
    \centering
    \includegraphics[width=1.0\linewidth]{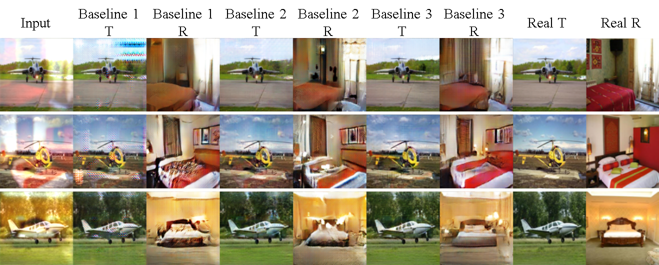}
    \caption{
    Examples of reflection separation results using baseline networks in Figure~\ref{fig:nets}.
    Best viewed in color with a digital zoom.
    }
\label{fig:compare_baseline}
\end{figure}

\vspace{2pt}
\noindent {\bf Network design for synthetic image reflection separation.}
The first baseline model simply maps an input image into two domains using an encoder and two domain-specific decoders based on the U-net architecture \cite{ronneberger2015u} as shown in Figure~\ref{fig:nets}(a).
The loss function for this network is an extension of \cite{isola2017image} as follows:
\begin{equation}\label{eq:loss1}
\begin{aligned}
  & \mathcal{L}_1(G_t, G_r, D_t, D_r) = \mathcal{L}_{adv}(G_t, D_t) + \mathcal{L}_{adv}(G_r, D_r) \\
  & \quad + \lambda_1\sum_{\alpha\in\{t,r\}}\mathbb{E}_{y,\alpha\sim p_{data}(y,\alpha)}[\|\alpha-G_\alpha(y)\|_1],
\end{aligned}
\end{equation}
where $\lambda_1$ controls relative importance of objectives.
It asks generators to not only fool discriminators but also to be similar to ground truth images.
However, this simple extension renders numerous artifacts and fails to separate two scenes suitably, as shown in Figure~\ref{fig:compare_baseline}.
We attribute this to a lack of communication between two generation branches.
As two images should be generated jointly in our problem, we need extra channels to communicate with each other.

To address this issue, we add a reconstruction branch, $G_y$, and share weights of the first few layers with generation branches as shown in Figure~\ref{fig:nets}(b).
In this case, the loss function is:
\begin{equation}\label{eq:loss2}
\begin{aligned}
  & \mathcal{L}_2(G_t, G_r, G_y, D_t, D_r) = \mathcal{L}_{adv}(G_t, D_t) + \mathcal{L}_{adv}(G_r, D_r) \\
  & \quad + \lambda_1\sum_{\alpha\in\{y,t,r\}}\mathbb{E}_{y,\alpha\sim p_{data}(y,\alpha)}[\|\alpha-G_\alpha(y)\|_1].
\end{aligned}
\end{equation}
%
Note that this is different from the weight sharing scheme of \cite{liu2016coupled}:
early weights of two generation branches are shared (i.e., semantics between two domains are shared).
On the other hand, in our problem, the semantics are often not shared between a transmitted scene and a reflected scene.
Instead, both scenes share semantics with an observed image.
Thus, by putting a reconstruction branch in the middle, all networks can communicate with each other during generation.
This approach helps decrease artifacts and separates the scene better as shown in Figure~\ref{fig:compare_baseline}.

The third model is shown in Figure~\ref{fig:nets}(c).
It aims to make use of high-level information in addition to the pixel-level appearance to assess generated images.
In this paper, we minimize the difference of contents between generated images and the input image.
A straightforward solution is to put a constraint that generated images should reconstruct an input image as faithfully as possible using the synthesis model.
%
However, it is a limited approach since the real synthesis models are unknown or non-differentiable, \eg,~\cite{fan2017generic}.
To address this issue, we compare feature maps of the input and generated images.
In Figure~\ref{fig:nets}(c), we introduce a new variable $w_y$ which estimates
the ratio of contents between a transmitted scene and reflected scene in $y$.
Using this parameter, a content loss between the observed scene and generated scenes is defined as follows:
\begin{equation}\label{eq:content_loss}
\begin{aligned}
  & \mathcal{L}_{content}(G_t, G_r) = \\
  & \sum_i \frac{1}{V_i}\Big(\|f_i(y) - (w_y f_i(G_t(y)) + (1-w_y)f_i(G_r(y)))\|_2 \\
  & \qquad\ + \|f_i(t)-f_i(G_t(y))\|_2 + \|f_i(r)-f_i(G_r(y))\|_2 \Big),
\end{aligned}
\end{equation}
where $f_i$ and $V_i$ denote a feature map and a volume of the $i$-th layer of the encoder, respectively.
It allows us to train the network for arbitrary synthesis models.
The loss for the network is defined as follows:
\begin{equation}\label{eq:loss}
\begin{aligned}
  & \mathcal{L}(G_t, G_r, G_y, D_t, D_r) = \\
  & \quad \mathcal{L}_{adv}(G_t, D_t) + \mathcal{L}_{adv}(G_r, D_r) + \\
  & \quad \lambda_1\sum_{\alpha\in\{y,t,r\}}\mathbb{E}_{y,\alpha\sim p_{data}(y,\alpha)}[\|\alpha-G_\alpha(y)\|_1] + \\
  & \quad \lambda_2\mathcal{L}_{content}(G_t, G_r),
\end{aligned}
\end{equation}
where $\lambda_1=\lambda_2=100$ are used for all experiments.

\subsection{Reflection Separation using Real Images}
\vspace{-2mm}
In this section, we describe how to train a network using real observations without ground truth images of the transmitted scene and reflected scene.
Note that the discussed networks so far rely on synthetic images since it is difficult to obtain ground truths of corresponding $t$ and $r$.
Even in the recent attempt for data collection \cite{wan2017benchmarking},
the original scene before reflection cannot be obtained.
It limits the performance of reflection separation due to the difference between the real observation and synthesized images.

\begin{figure}[t]
    \centering
    \includegraphics[width=1.0\linewidth]{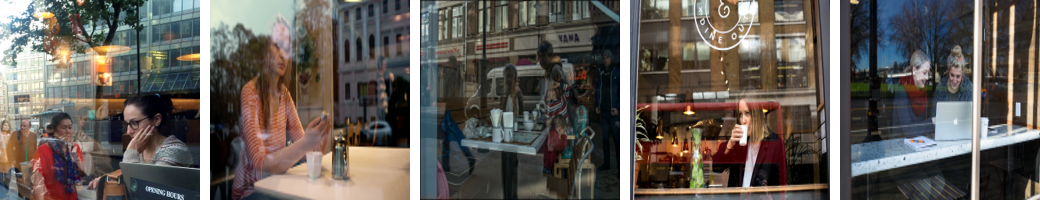}
    \caption{
    Examples of cafe images taken from the outside.
    Part of the cafe interior is not visible due to the reflection.
    }
\label{fig:ex_real}
\end{figure}
One of the largest gaps between real and synthesized images
is due to the combination weight $w$ in the synthesis model.
For real images, reflected scenes are often dominant in certain regions of an observed image
where transmitted signals are unrecognizable and can only be guessed.
For example, as shown in Figure~\ref{fig:ex_real}, cafe interiors are not visible due to reflected buildings.
However, state-of-the-art synthesis methods use a scalar $w$ to combine two scenes.
To alleviate this issue,
we predict a mask map using a new branch $G_m$ instead of a scalar $w_y$ as shown in Figure~\ref{fig:nets}(d).
For each location, its value is between $[0,1]$ which represents a confidence score that the pixel belongs to a transmitted scene.
As such, we have $G_m(y)\otimes y \simeq G_m(y)\otimes G_t(y) = G_{mt}(y)$ and
$(1-G_m(y))\otimes y \simeq (1-G_m(y))\otimes G_r(y) = G_{mr}(y)$,
where $\otimes$ denotes pixel-wise multiplication and $G_{mt}$ and $G_{mr}$ are defined for notational simplicity.
The confidence map prediction branch and other generation branches are trained iteratively.

The network can be trained in a weakly supervised manner, i.e.,
tuples of $(y,t,r)$ are given where $t$ and $r$ are images belonging to the same categories of transmitted and reflected scenes included in $y$, respectively.
In this case, $t$ and $r$ are not conditioned on $y$.
Thus, (\ref{eq:cgan_loss}) is changed to:
\begin{equation}\label{eq:gan_loss}
\begin{aligned}
    \mathcal{L}_{adv}(G_t,D_t) =
    & \mathbb{E}_{t\sim p_{data}(t)}[\log D_t(t)] + \\
    & \mathbb{E}_{y\sim p_{data}(y)}[\log (1-D_t(G_t(y))].
\end{aligned}
\end{equation}
%
%
In addition, a loss function in (\ref{eq:loss}) should be changed for weakly supervised learning.
By combining all the losses, the overall loss function becomes
\begin{equation}\label{eq:loss_unsup} \small
\begin{aligned}
  & \mathcal{L}(G_t, G_r, G_y, G_m, D_t, D_r) \\
  & = \mathcal{L}_{adv}(G_t, D_t) + \mathcal{L}_{adv}(G_r, D_r) \\
  & + \lambda_1\mathbb{E}_{y\sim p_{data}(y)}[\|y-G_y(y)\|_2] + \|G_m(y)\otimes y -G_{mt}\|_2 + \\
  & \qquad\qquad\qquad\qquad \|(1-G_m(y))\otimes y - G_{mr}(y)\|_2] \\
  & + \lambda_2\sum_i \frac{1}{V_i}\big(\|f_i(G_m(y)\otimes y) - f_i(G_{mt}(y))\|_2 + \\
  & \qquad\qquad\qquad \|f_i((1-G_m(y))\otimes y) - f_i(G_{mr}(y))\|_2\big).
\end{aligned}
\end{equation}
It only uses $t$ and $r$ to compute the adversarial loss which does not need the exact ground truth data of $y$.
Note that it allows us to learn reflection separation problem without any approximations for synthetic images.

\vspace{-2mm}
\section{Experimental Results} \label{sec:exp}
\vspace{-2mm}
\begin{figure}[t]
    \centering
    \includegraphics[width=1.0\linewidth]{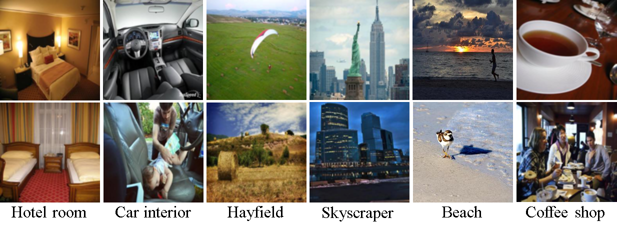}
    \caption{
    Examples of the Places dataset.
    It has a large number of categories and images in the same category are fairly diverse.
    }
\label{fig:places_dataset}
\end{figure}
We first describe the experimental settings.
We use the Places dataset which contains 8 million images of 365 scenes to synthesize images.
As shown in Figure~\ref{fig:places_dataset}, it has diverse images that cover large variations in the lighting condition, viewpoint, distance to the scene, and a number of objects in the scene.
For each experiment, two scenes are selected to synthesize training images.
For real images, we collect 178 photos of a cafe taken from the outside, from the Internet.
As shown in Figure~\ref{fig:ex_real}, the images contain challenging reflections.

\begin{figure*}[!t]
    \captionsetup[subfloat]{farskip=2pt,captionskip=1pt}
    \centering
    \subfloat[Input $y$]{\includegraphics[width=0.1\linewidth]{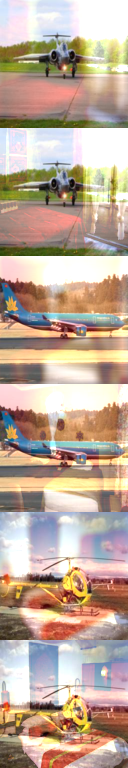}%
    \label{fig:syn_input}}
    \hspace{0.1pt}
    \subfloat[$G_t(y)$]{\includegraphics[width=0.1\linewidth]{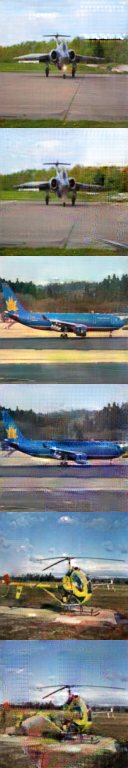}%
    \label{fig:syn_g1}}
    \hspace{0.1pt}
    \subfloat[$G_r(y)$]{\includegraphics[width=0.1\linewidth]{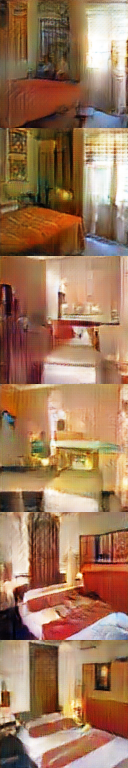}%
    \label{fig:syn_g2}}
    \hspace{0.1pt}
    \subfloat[\cite{arvanitopoulos2017single} $t$]{\includegraphics[width=0.1\linewidth]{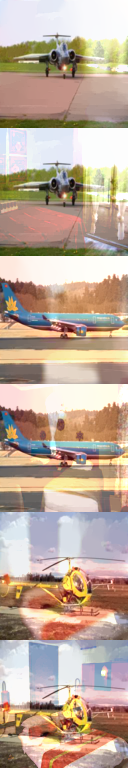}%
    \label{fig:syn_cvpr}}
    \hspace{0.1pt}
    \subfloat[\cite{fan2017generic} $t$]{\includegraphics[width=0.1\linewidth]{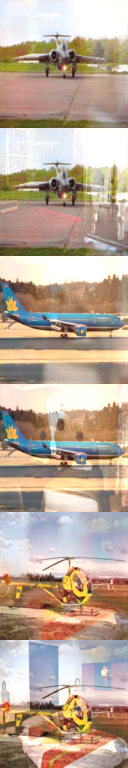}%
    \label{fig:syn_iccv1}}
    \hspace{0.1pt}
    \subfloat[\cite{fan2017generic} $r$]{\includegraphics[width=0.1\linewidth]{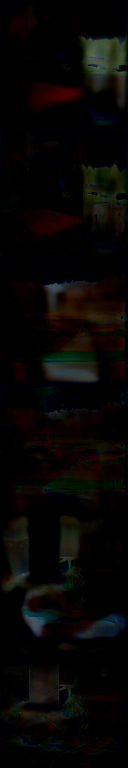}%
    \label{fig:syn_iccv2}}
    \hspace{0.1pt}
    \subfloat[Real $t$]{\includegraphics[width=0.1\linewidth]{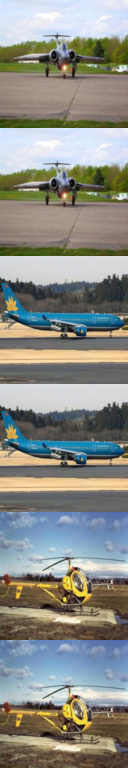}%
    \label{fig:syn_gt1}}
    \hspace{0.1pt}
    \subfloat[Real $r$]{\includegraphics[width=0.1\linewidth]{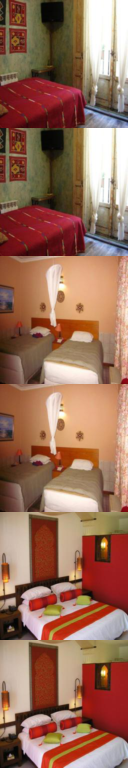}%
    \label{fig:syn_gt2}}
    \hfil
    \caption{
    Single image reflection separation results on synthetic images of the Places dataset.
    In this experiment, images are synthesized based on the clipping model \cite{fan2017generic}.
    For every two rows, we show results with and without the blurry assumption for the same input pair.
    }
    \label{fig:syn_results}
\end{figure*}

\begin{table*}[t]
    \centering
    \footnotesize
    \caption{
    Quantitative results of single image reflection separation on the Places dataset.
    For comparison, we run experiments using each author's publicly available implementation.
    Note that \cite{shih2015reflection} does not work on other physical models and \cite{arvanitopoulos2017single} does not provide a reflection scene as an output.
    }
    \begin{tabular}{ccc|c|c|c|c|c|c|c|c|c|c|}
    \cline{4-13}
    & & & \multicolumn{5}{ c| }{Transmission} & \multicolumn{5}{ c| }{Reflection} \\ \cline{4-13}
    Dataset& Physical model & & \cite{li2014single} & \cite{shih2015reflection} & \cite{arvanitopoulos2017single} & \cite{fan2017generic} & Ours & \cite{li2014single} & \cite{shih2015reflection} & \cite{arvanitopoulos2017single} & \cite{fan2017generic} & Ours \\ \cline{1-13}
    \multicolumn{1}{ |c  }{\multirow{10}{*}{\shortstack{$t$: Airfield,\\ $r$: Hotel room}} } &
    \multicolumn{1}{ |c  }{\multirow{2}{*}{\cite{shih2015reflection}} } & 
    \multicolumn{1}{ |c|  }{PSNR} & 14.4 & 13.1 & 17.3 & 17.4 & \textbf{18.2} & 10.1 & 10.2 & - & 7.28 & \textbf{18.9} \\
    \multicolumn{1}{ |c  }{} & \multicolumn{1}{ |c  }{} &
    \multicolumn{1}{ |c|  }{SSIM} & 0.34 & 0.26 & 0.42 & 0.44 & \textbf{0.45} & 0.10 & 0.14 & - & 0.06 & \textbf{0.48} \\ \cline{2-13}
    \multicolumn{1}{ |c  }{} &
    \multicolumn{1}{ |c  }{\multirow{2}{*}{\cite{arvanitopoulos2017single}} } & 
    \multicolumn{1}{ |c|  }{PSNR} & 15.8 & - & 19.9 & 20.1 & \textbf{26.3} & 11.8 & - & - & 8.0 & \textbf{18.0} \\
    \multicolumn{1}{ |c  }{} & \multicolumn{1}{ |c  }{} &
    \multicolumn{1}{ |c|  }{SSIM} & 0.47 & - & 0.58 & 0.61 & \textbf{0.80} & 0.17 & - & - & 0.11 & \textbf{0.37} \\ \cline{2-13}
    \multicolumn{1}{ |c  }{} &
    \multicolumn{1}{ |c  }{\multirow{2}{*}{\shortstack{\cite{arvanitopoulos2017single} without \\ blur assumption}} } & 
    \multicolumn{1}{ |c|  }{PSNR} & 13.8 & - & 17.4 & 17.4 & \textbf{21.5} & 10.0 & - & - & 7.2 & \textbf{17.5} \\
    \multicolumn{1}{ |c  }{} & \multicolumn{1}{ |c  }{} &
    \multicolumn{1}{ |c|  }{SSIM} & 0.31 & - & 0.41 & 0.42 & \textbf{0.63} & 0.10 & - & - & 0.05 & \textbf{0.45} \\ \cline{2-13}
    \multicolumn{1}{ |c  }{} &
    \multicolumn{1}{ |c  }{\multirow{2}{*}{\cite{fan2017generic}} } & 
    \multicolumn{1}{ |c|  }{PSNR} & 15.1 & - & 15.9 & 18.1 & \textbf{26.1} & 9.95 & - & - & 8.1 & \textbf{17.5} \\
    \multicolumn{1}{ |c  }{} & \multicolumn{1}{ |c  }{} &
    \multicolumn{1}{ |c|  }{SSIM} & 0.49 & - & 0.53 & 0.59 & \textbf{0.81} & 0.10 & - & - & 0.12 & \textbf{0.35} \\ \cline{2-13}
    \multicolumn{1}{ |c  }{} &
    \multicolumn{1}{ |c  }{\multirow{2}{*}{\shortstack{\cite{fan2017generic} without \\ blur assumption}} } & 
    \multicolumn{1}{ |c|  }{PSNR} & 13.6 & - & 15.1 & 15.6 & \textbf{22.8} & 9.67 & - & - & 7.3 & \textbf{17.5} \\
    \multicolumn{1}{ |c  }{} & \multicolumn{1}{ |c  }{} &
    \multicolumn{1}{ |c|  }{SSIM} & 0.40 & - & 0.49 & 0.50 & \textbf{0.68} & 0.09 & - & - & 0.11 & \textbf{0.41} \\ \cline{1-13}
    \multicolumn{1}{ |c  }{\multirow{10}{*}{\shortstack{$t$: Skyscraper,\\ $r$: Conference room}} } &
    \multicolumn{1}{ |c  }{\multirow{2}{*}{\cite{shih2015reflection}} } & 
    \multicolumn{1}{ |c|  }{PSNR} & 14.5 & 14.5 & 17.6 & 17.4 & \textbf{21.1} & 9.80 & 10.8 & - & 7.30 & \textbf{16.5} \\
    \multicolumn{1}{ |c  }{} & \multicolumn{1}{ |c  }{} &
    \multicolumn{1}{ |c|  }{SSIM} & 0.39 & 0.34 & 0.45 & 0.46 & \textbf{0.62} & 0.10 & 0.11 & - & 0.02 & \textbf{0.37} \\ \cline{2-13}
    \multicolumn{1}{ |c  }{} &
    \multicolumn{1}{ |c  }{\multirow{2}{*}{\cite{arvanitopoulos2017single}} } & 
    \multicolumn{1}{ |c|  }{PSNR} & 15.3 & - & 19.2 & 19.9 & \textbf{24.5} & 10.9 & - & - & 8.4 & \textbf{16.4} \\
    \multicolumn{1}{ |c  }{} & \multicolumn{1}{ |c  }{} &
    \multicolumn{1}{ |c|  }{SSIM} & 0.48 & - & 0.60 & 0.64 & \textbf{0.76} & 0.11 & - & - & 0.10 & \textbf{0.31} \\ \cline{2-13}
    \multicolumn{1}{ |c  }{} &
    \multicolumn{1}{ |c  }{\multirow{2}{*}{\shortstack{\cite{arvanitopoulos2017single} without \\ blur assumption}} } & 
    \multicolumn{1}{ |c|  }{PSNR} & 14.3 & - & 17.0 & 16.7 & \textbf{20.4} & 9.70 & - & - & 7.1 & \textbf{16.8} \\
    \multicolumn{1}{ |c  }{} & \multicolumn{1}{ |c  }{} &
    \multicolumn{1}{ |c|  }{SSIM} & 0.35 & - & 0.42 & 0.41 & \textbf{0.57} & 0.08 & - & - & 0.01 & \textbf{0.40} \\ \cline{2-13}
    \multicolumn{1}{ |c  }{} &
    \multicolumn{1}{ |c  }{\multirow{2}{*}{\cite{fan2017generic}} } & 
    \multicolumn{1}{ |c|  }{PSNR} & 14.6 & - & 15.3 & 17.2 & \textbf{23.1} & 10.0 & - & - & 8.1 & \textbf{15.7} \\
    \multicolumn{1}{ |c  }{} & \multicolumn{1}{ |c  }{} &
    \multicolumn{1}{ |c|  }{SSIM} & 0.46 & - & 0.51 & 0.58 & \textbf{0.73} & 0.07 & - & - & 0.09 & \textbf{0.28} \\ \cline{2-13}
    \multicolumn{1}{ |c  }{} &
    \multicolumn{1}{ |c  }{\multirow{2}{*}{\shortstack{\cite{fan2017generic} without \\ blur assumption}} } & 
    \multicolumn{1}{ |c|  }{PSNR} & 12.9 & - & 14.5 & 15.6 & \textbf{20.7} & 9.24 & - & - & 7.3 & \textbf{16.3} \\
    \multicolumn{1}{ |c  }{} & \multicolumn{1}{ |c  }{} &
    \multicolumn{1}{ |c|  }{SSIM} & 0.37 & - & 0.45 & 0.50 & \textbf{0.62} & 0.07 & - & - & 0.11 & \textbf{0.37} \\ \cline{1-13}
    \end{tabular}
    \label{tab:psnr_ssim}
\end{table*}
\begin{figure}[!t]
    \captionsetup[subfloat]{farskip=2pt,captionskip=1pt}
    \centering
    \subfloat[Airfield and beach]{\includegraphics[width=0.8\linewidth]{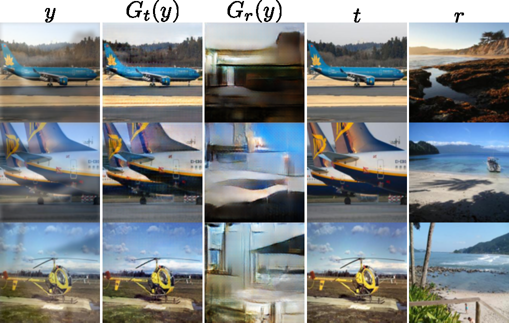}%
    \label{fig:test_ac}} \\
    \subfloat[Beach and street]{\includegraphics[width=0.8\linewidth]{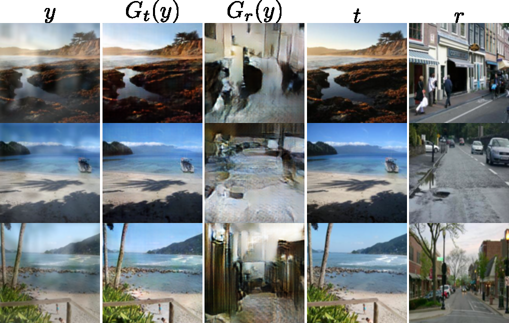}%
    \label{fig:test_cd}} \\
    \subfloat[Beach and hotel room]{\includegraphics[width=0.8\linewidth]{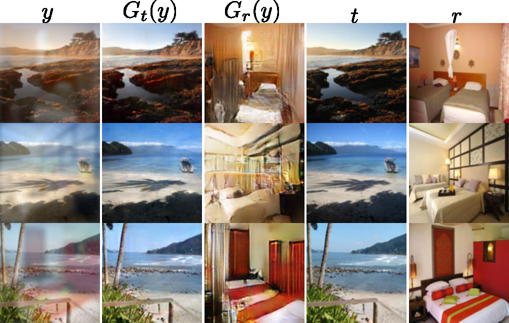}%
    \label{fig:test_cb}} \\
    \subfloat[Sky and hotel room]{\includegraphics[width=0.8\linewidth]{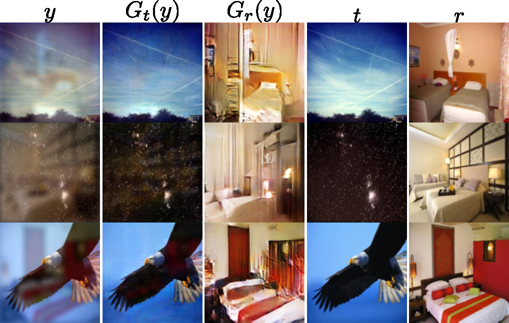}%
    \label{fig:test_db}}
    \caption{
    The proposed network is trained on airfield and hotel room and then tested on other categories.
    Other categories are completely not fed to the network during training.
    }
    \label{fig:test_other}
\end{figure}
For training, networks shown in Figure~\ref{fig:nets}(b), Figure~\ref{fig:nets}(c), and Figure~\ref{fig:nets}(d) share weights of the first three layers in the generator.
For each mini-batch, we first resize images into $256\times256$ pixels.
A rectangular patch is cropped with a height and a width sampled from $[192, 256]$ uniformly at random.
Then, it is resized to $128\times128$ pixels for training.
For the blurring model, we pick $w\in[0.5,0.7]$ and $\sigma$ of the Gaussian kernel between $[2,5]$.
%
More quantitative and qualitative results are presented in the supplementary material.
The source code will be made available to the public.
%

\subsection{Synthetic Images}
\vspace{-2mm}
In this section, we show reflection separation results using a network trained on synthetic data.
We evaluate the proposed algorithm against the state-of-the-art methods \cite{fan2017generic, arvanitopoulos2017single, shih2015reflection}.
As they use different approximation methods to model the reflection,
we provide results for all cases.

Figure \ref{fig:syn_results} shows reflection separation results when the clipping model \cite{fan2017generic} is used to synthesize images.
We alternatively use the blur assumption for each row.
In most cases, the proposed network in Figure~\ref{fig:nets}(c) successfully removes reflection artifacts on the transmitted scene and recovers the reflected scene reasonably.
In contrast, other methods generate
unclear images of transmitted scenes and barely recognizable reflected scenes.
When the blurry assumption is not used, all other methods fail to suppress reflections.
In addition, they are sensitive to the synthesis model.
The results show that other methods are designed specifically based on their assumptions.
On the other hand, the proposed algorithm performs favorably in all cases.
Table~\ref{tab:psnr_ssim} shows that the proposed algorithm performs
favorably against other methods quantitatively.

A network trained with two categories is evaluated for different types of scenes.
Figure~\ref{fig:test_other} shows the results when the proposed network is trained on airfield and hotel room categories
and then tested on various scenes using the blurring model \cite{arvanitopoulos2017single}.
When a category of the reflected scene is changed as shown in Figure~\ref{fig:test_other}(a),
or both are changed as shown in Figure~\ref{fig:test_other}(b),
the transmitted scene is well restored while the reflected scene is not realistic.
On the other hand, when the transmitted scene is changed as shown in Figure~\ref{fig:test_other}(c) and Figure~\ref{fig:test_other}(d),
both scenes are recovered properly.
These results indicate that the transmission branch learns how to remove blurry regions and generate the image based on the context of sharp regions
while the reflection branch focuses on recovering the original scene from a blurry observation.
However, as presented in the supplementary material, the network is not able to separate reflections on real images.
%
%
The state-of-the-art methods based on the same synthesis models also fail to separate reflections.
The findings are consistent with the observation in \cite{wan2017benchmarking} that existing methods do not perform well on real images.
In addition, we put randomness while synthesizing training images to increase the generalization ability of the network as mentioned before.
We attribute this failure to a non-realistic synthesis model.
The separation performance can be improved when a realistic synthesis model is given since the proposed network can be trained with any synthesis methods.

\begin{figure}[t]
    \captionsetup[subfloat]{farskip=2pt,captionskip=1pt}
    \centering
    \includegraphics[width=0.9\linewidth]{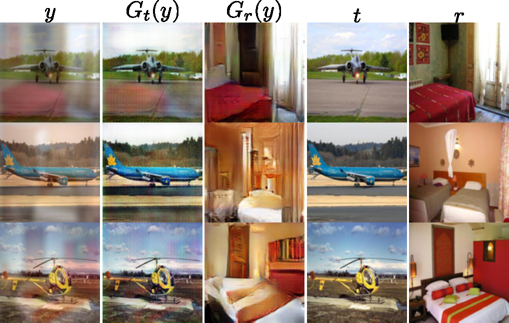}
    \caption{Sample results from unsupervised reflection separation for synthetic images.
    }
\label{fig:unsup}
\end{figure}
\begin{figure*}[!t]
    \captionsetup[subfloat]{farskip=2pt,captionskip=1pt}
    \centering
    \subfloat[Input $y$]{\includegraphics[width=0.1\linewidth]{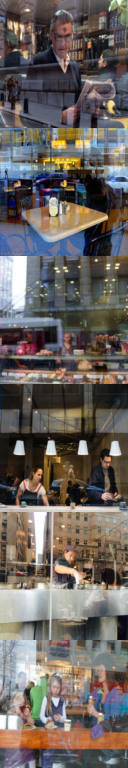}%
    \label{fig:real_input}}
    \hspace{0.1pt}
    \subfloat[\cite{arvanitopoulos2017single} $t$]{\includegraphics[width=0.1\linewidth]{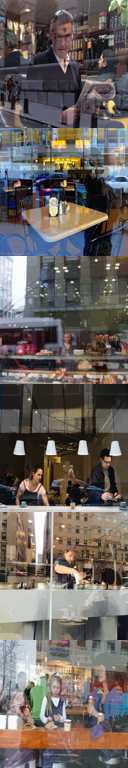}%
    \label{fig:real_cvpr}}
    \hspace{0.1pt}
    \subfloat[\cite{fan2017generic} $t$]{\includegraphics[width=0.1\linewidth]{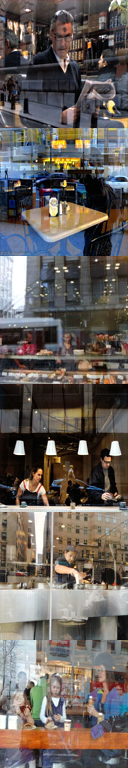}%
    \label{fig:real_iccv1}}
    \hspace{0.1pt}
    \subfloat[\cite{fan2017generic} $r$]{\includegraphics[width=0.1\linewidth]{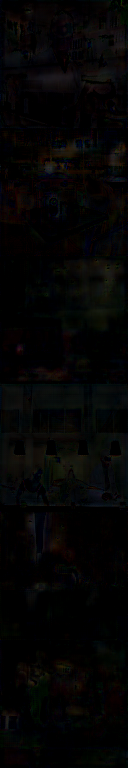}%
    \label{fig:real_iccv2}}
    \hspace{0.1pt}
    \subfloat[$G_t(y)$]{\includegraphics[width=0.1\linewidth]{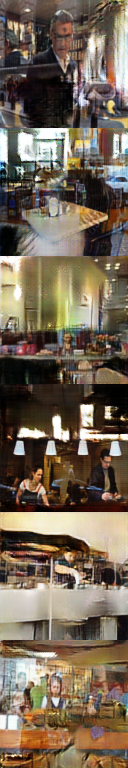}%
    \label{fig:real_g1}}
    \hspace{0.1pt}
    \subfloat[$G_r(y)$]{\includegraphics[width=0.1\linewidth]{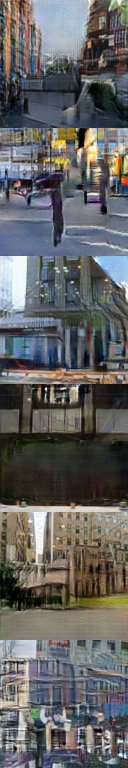}%
    \label{fig:real_g2}}
    \hspace{0.1pt}
    \subfloat[$G_m(y)$]{\includegraphics[width=0.1\linewidth]{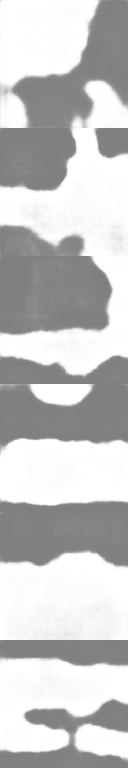}%
    \label{fig:real_mask}}
    \hspace{0.1pt}
    \subfloat[$G_{mt}(y)$]{\includegraphics[width=0.1\linewidth]{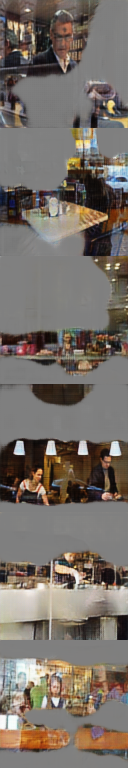}%
    \label{fig:real_g1mask}}
    \hspace{0.1pt}
    \subfloat[$G_{mr}(y)$]{\includegraphics[width=0.1\linewidth]{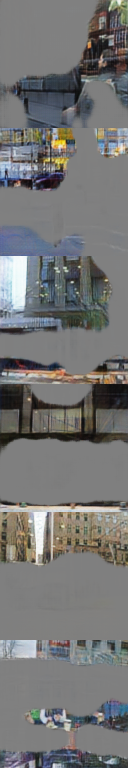}%
    \label{fig:real_g2mask}}
    \hfil
    \caption{
    Single image reflection separation results on real images.
    }
    \label{fig:real_results}
\end{figure*}
\begin{figure}[!t]
    \captionsetup[subfloat]{farskip=2pt,captionskip=1pt}
    \captionsetup[subfloat]{font=scriptsize,labelfont=scriptsize}
    \centering
    \subfloat[Input $y$]{\includegraphics[width=0.155\linewidth]{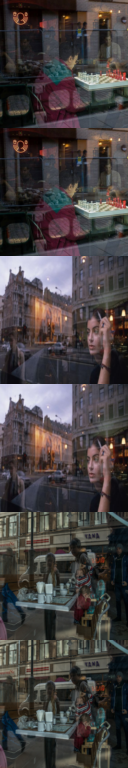}%
    \label{fig:real_input}}
    \hspace{0.1pt}
    \subfloat[$G_t(y)$]{\includegraphics[width=0.155\linewidth]{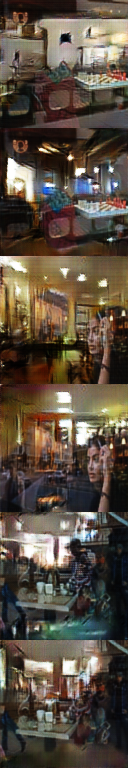}%
    \label{fig:real_g1}}
    \hspace{0.1pt}
    \subfloat[$G_r(y)$]{\includegraphics[width=0.155\linewidth]{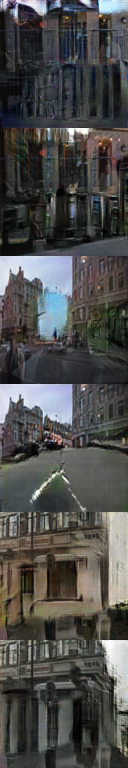}%
    \label{fig:real_g2}}
    \hspace{0.1pt}
    \subfloat[$G_m(y)$]{\includegraphics[width=0.155\linewidth]{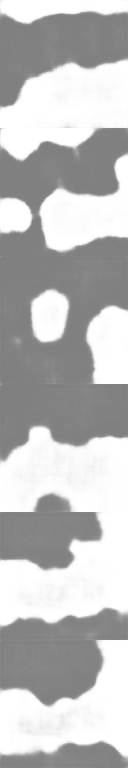}%
    \label{fig:real_mask}}
    \hspace{0.1pt}
    \subfloat[$G_{mt}(y)$]{\includegraphics[width=0.155\linewidth]{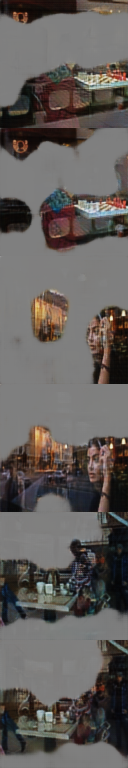}%
    \label{fig:real_g1mask}}
    \hspace{0.1pt}
    \subfloat[$G_{mr}(y)$]{\includegraphics[width=0.155\linewidth]{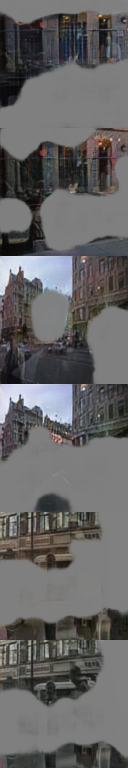}%
    \label{fig:real_g2mask}}
    \hfil
    \caption{
    Different separation results for the same input image.
    }
    \label{fig:real_results_sim}
\end{figure}
For a weakly supervised reflection separation,
the training data is split into two halves.
The first half is used to synthesize $y$ and the second half is fed to the discriminator as real images for training.
Figure~\ref{fig:unsup} shows that
the proposed algorithm separates a synthesized scene well without any ground truth images of transmission and reflection.

\vspace{-1mm}
\subsection{Real Images}
\vspace{-2mm}
Figure~\ref{fig:real_results} shows results corresponding to the network in Figure~\ref{fig:nets}(d)
trained with real data in a weakly supervised manner.
Input images contain not only sharp or blurry reflections but also spatially non-uniform reflection regions.
For example, at the second row of Figure~\ref{fig:real_results}(a), most parts of cafe interiors are not visible.
If reflections are simply suppressed, then the remaining transmitted scene would be dark and less informative.
%
Moreover, it is incorrect to return a dark transmitted scene in this case since we can observe that lights of the cafe are turned on.
Therefore, it is challenging to separate real input images as we need to infer invisible regions at the same time.

As the issue is not handled in state-of-the-art methods, they rarely suppress or separate reflections as shown in Figure~\ref{fig:real_results}(b), Figure~\ref{fig:real_results}(c) and Figure~\ref{fig:real_results}(d).
On the other hand, the proposed network can decompose an input image into two reasonable scenes.
Figure~\ref{fig:real_results}(e) and Figure~\ref{fig:real_results}(f) show restored transmitted scene and reflected scene.
The confidence map of a transmitted scene is shown in Figure~\ref{fig:real_results}(g).
Figure~\ref{fig:real_results}(h) and Figure~\ref{fig:real_results}(i) are masked scenes for better visualization of the confidence map.
%

Single image reflection separation is an underdetermined problem which has many solution candidates.
In Figure~\ref{fig:real_results_sim}, we show separation results for the same input image using two different random seeds.
For each pair, while networks capture similar parts for realistic separations,
generated images are different regarding to structure and appearance of the scene.

\vspace{-2mm}
\section{Conclusions}
\vspace{-2mm}
We propose an algorithm for single image reflection separation problem.
As it is an ill-posed problem, we assume that categories of transmission and reflection scenes are known.
It allows us to remove conventional assumptions, such as the blurry reflected scene, that are not realistic in many cases.
We design convolutional neural networks based on adversarial losses to separate an observed image into the transmitted scene and reflected scene by generating them.
Experimental results show that the proposed algorithm performs favorably against the state-of-the-art methods,
particularly for recovering reflected scenes.
%
For synthetic images, the transmitted scene is reliably restored without knowing the type of two scenes.
In addition, we demonstrate that the network can be trained in a weakly-supervised manner,
i.e., the network is trained on real images only.

{\small
\bibliographystyle{ieee}
\bibliography{cvpr2018_reflection}
}

\end{document}